\documentclass[letterpaper, 10 pt, journal, twoside]{IEEEtran}

\usepackage{times}
\usepackage{mathtools}
\usepackage{amsfonts}
\usepackage{amsmath}
\usepackage{amssymb}
\usepackage{flafter}
\usepackage{todonotes}
\usepackage{cleveref}
\usepackage{graphicx}
\usepackage{wrapfig}
\usepackage{graphics}
\usepackage{epsfig}
\usepackage[font=small,tableposition=top]{caption}
\usepackage[ruled,vlined,linesnumbered,noend]{algorithm2e}
\usepackage[noend]{algorithmic}
\usepackage{multirow}
\usepackage{units}
\usepackage{cite}
\usepackage{etoolbox}
\usepackage{hyperref}
\usepackage{url}
\usepackage[normalem]{ulem}

\makeatletter
\patchcmd{\@makecaption}
  {\scshape}
  {}
  {}
  {}
\makeatother
\newcommand{\norm}[1]{\left\lVert#1\right\rVert}

\title{DeepGait: Planning and Control of Quadrupedal Gaits using Deep Reinforcement Learning}

\author{Vassilios Tsounis$^{*}$, Mitja Alge$^{*}$, Joonho Lee, Farbod Farshidian, and  Marco Hutter%
\thanks{Manuscript received: September, 10, 2019; Revised December, 17, 2019; Accepted January, 18, 2020.}%
\thanks{This paper was recommended for publication by Editor Nikolaos G. Tsagarakis upon evaluation of the Associate Editor and Reviewers' comments.}%
\thanks{This work was supported by Intel Labs, the Swiss National Science Foundation (SNSF) through project 166232, 188596, the National Centre of Competence in Research Robotics (NCCR Robotics), and the European Union's Horizon 2020 research and innovation program under grant agreement No.780883. Moreover, this work has been conducted as part of ANYmal Research, a community to advance legged robotics.}%
\thanks{All authors are with the Robotic Systems Lab, ETH Z\"u{}rich, Switzerland. {\tt\footnotesize tsounisv@ethz.ch}}%
\thanks{$^{*}$\,These authors contributed equally.}%
\thanks{Digital Object Identifier (DOI): see top of this page.}
}

\begin{document}
\maketitle

\begin{abstract}
This paper addresses the problem of legged locomotion in non-flat terrain. As legged robots such as quadrupeds are to be deployed in terrains with geometries which are difficult to model and predict, the need arises to equip them with the capability to generalize well to unforeseen situations. In this work, we propose a novel technique for training neural-network policies for terrain-aware locomotion, which combines state-of-the-art methods for model-based motion planning and reinforcement learning. Our approach is centered on formulating Markov decision processes using the evaluation of dynamic feasibility criteria in place of physical simulation. We thus employ policy-gradient methods to independently train policies which respectively plan and execute foothold and base motions in 3D environments using both proprioceptive and exteroceptive measurements. We apply our method within a challenging suite of simulated terrain scenarios which contain features such as narrow bridges, gaps and stepping-stones, and train policies which succeed in locomoting effectively in all cases.
\end{abstract}

\begin{IEEEkeywords}
Legged Robots; Deep Learning in Robotics and Automation; Motion and Path Planning
\end{IEEEkeywords}

\section{Introduction}
\label{sec:Introduction}
\begin{figure}[t]
    \centering
    \includegraphics[width=\linewidth]{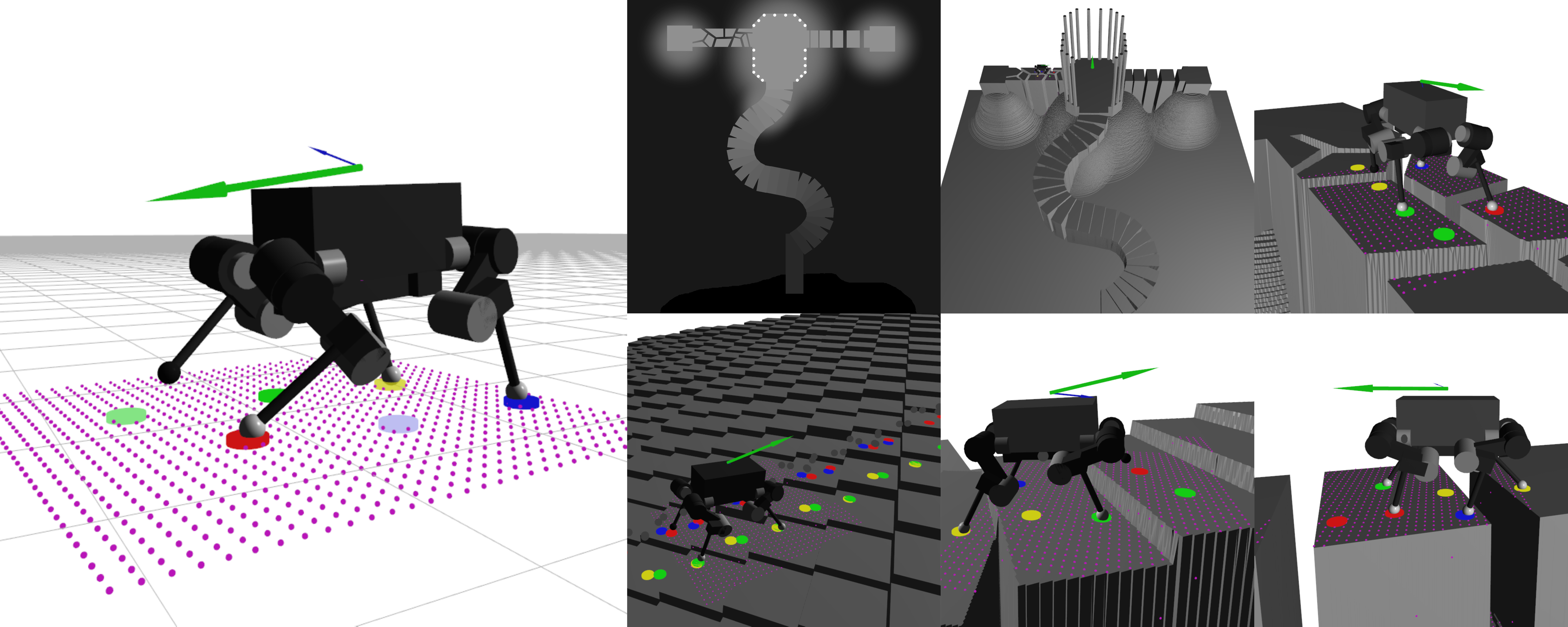}
    \caption{The suite of terrains: the baseline Flat-World scenario (left), the Random-Stairs scenario (bottom center), and composite Temple-Ascent (right) scenario comprising a set of winding stairs and two derelict bridges containing stepping-stones and gaps of varying size.}
    \label{fig:TerrainSuite}
\end{figure}
\IEEEPARstart{L}{egged} locomotion in non-flat terrain, both structured and unstructured, poses a significant challenge in robotics. Operating autonomously in such environments requires addressing the problem of multi-contact motion planning and control. If a legged robot such as ANYmal~\cite{hutter2016anymal} is to traverse complex environments autonomously, it must possess the capability to select footholds appropriate for the terrain,  while also retaining balance at all times. This work deals specifically with the problem of planning and executing sequences of footholds for quadrupedal locomotion in rigid non-flat terrain using proprioceptive and exteroceptive sensing.

Dynamically walking on non-flat terrain necessitates the optimization of continuous state-input trajectories such as the motion of the base, as well as discrete decision variables such as which surface, and when, to make contact with. This has been addressed predominantly using model-based approaches, such as those employing deterministic optimization techniques~\cite{kuindersma2016optimization,winkler2018gait}, in conjunction with other heuristics~\cite{fankhauser2018robust}, to plan motions for both the base and feet. Although some of the aforementioned approaches are able to solve such problems compromising both continuous and discrete variables~\cite{kuindersma2016optimization,winkler2018gait}, these remain too computationally intensive to be executed online. Thus, only kinostatic approaches~\cite{doi:10.1002/rob.21610,fankhauser2018robust} have managed to perform online foothold planning. Also, most methods typically employ some form of parameterization or qualification of the terrain~\cite{Kalakrishnan:2011:LPC:1936801.1936805,fankhauser2018robust,tonneau2018reachability} in order to simplify the search for viable footholds.

One of the primary challenges in multi-contact planning for multi-legged systems lies in dealing with the combinatorial problem due to the vast number of contact configurations admissible for the terrain. Typical solutions involve either assuming the gait pattern~\cite{doi:10.1002/rob.21610,fankhauser2018robust} or employing sampling-based search techniques~\cite{tonneau2018efficient,tonneau2018reachability}. There also exist works that have combined both optimization and sampling-based methods~\cite{Kalakrishnan:2011:LPC:1936801.1936805,doi:10.1002/rob.21610,griffin2019footstep}. However, these typically resort to decoupling the selection of footholds from the optimization of base motions and thus remain kinostatic as they tend to neglect the dynamics of the system.

Some works have also incorporated machine-learning techniques for facilitating terrain perception~\cite{Kalakrishnan:2011:LPC:1936801.1936805,magana2019fast,klamt2019towards}. Others, have employed Deep Reinforcement Learning (DRL)~\cite{heess2017emergence,Peng:2017:DDL:3072959.3073602} for realizing end-to-end terrain-aware locomotion. The use of the latter, however, still poses several challenges, namely: (1) how to eliminate undesirable yet retain beneficial emergent behavior, and (2) reduce overall sample complexity and train policies efficiently.

We propose a new method that combines state-of-the-art model-based and model-free methods to enable quadrupedal systems to traverse complex non-flat terrain. Our formulation consists of: (1) a terrain-aware planner that generates sequences of footholds and base motions that direct the robot towards a target heading, and (2) a foothold and base motion controller which executes the aforementioned sequence while maintaining balance and dealing with disturbances. Both planner and controller are realized as stochastic policies parameterized using Neural-Network (NN) function approximation, which are optimized using state-of-the-art Deep Reinforcement Learning (DRL) algorithms.

Our contributions with this work are:
(1) A novel method for training kinodynamic motion planners, which employs a Trajectory Optimization (TO) technique for determining so-called \textit{transition feasibility} between discrete support phases using a coarse model of the robot. This removes the need for a planner to interact with both physics and a controller during training, allows the two policies to be trained independently, and leads to a significant reduction in overall sample complexity.
(2) A simple formulation for realizing dynamic walking controllers that use target footholds as references and rely solely on proprioceptive sensing. This allows us to train controllers that can fully exploit the kinematics and dynamics of the robot in order to track arbitrary target footholds, irrespective of the planner used to generate them.

We evaluate the performance of our method across a set of challenging locomotion scenarios using a physics simulator and present results thereof. Our experiments demonstrate that the planner can generalize well across terrain types, and the controller succeeds in tracking reference footholds while always balancing the robot. Moreover, we illustrate the advantages of our method by comparing it with a state-of-the-art model-based approach~\cite{fankhauser2018robust}.

\section{Preliminaries}
\label{sec:Preliminaries}

\subsection{Reinforcement Learning}
\label{subsec:ReinforcementLearning}
We consider the problem of sequential decision making in which an agent interacts with an environment with the objective of maximizing cumulative reward. We model this problem as a discrete-time infinite Markov Decision Processes (MDP) with a discounted expected return objective. Such an MDP consists of set of states $\mathcal{S}$, a set of actions $\mathcal{A}$, a transition dynamics distribution, an initial state distribution, a scalar reward function $r(\boldsymbol{s}_{t}, \boldsymbol{a}_{t}, \boldsymbol{s}_{t+1})$, and a scalar discount factor $\gamma \in [0,1)$. The agent selects actions according to a policy $\pi$ with the objective of maximizing the expected return $\mathbb{E}[\sum_{k=t}^{\infty} \gamma^{k}r_{t+k}]$, where $r_{t}$ is the scalar reward resulting from the state transition at time-step $t$. As we consider infinite MDPs in which $\mathcal{S}$ and $\mathcal{A}$ are infinite sets, we use parameterized stochastic policies $\pi_{\boldsymbol{\theta}}(\boldsymbol{a}|\boldsymbol{o}_{t})$, which are distributions over actions $\boldsymbol{a} \in \mathcal{A}$ conditioned on observations $\boldsymbol{o}_{t} \in \mathcal{O}$ given parameter vectors $\boldsymbol{\theta} \in \mathbb{R}^{n}$.

\subsection{Model of the System}
\label{subsec:SystemModel}

The robot comprises an unactuated floating base and four articulated legs with actuated rotational joints. The state of the robot is specified as: $\mathbf{r}_{B} \in \mathbb{R}^{3}$ the absolute\footnote{We define a global inertial frame $W$ for the world, and local body-fixed frames $B$ for the base and $F$ for the feet. Left sub-scripts denote the frame in which the vector is expressed, but omit it for absolute quantities.} position of the base, $\mathbf{R}_{B} \in SO(3)$ is the rotation matrix representing the absolute attitude of the base, $\mathbf{v}_{B},\boldsymbol{\omega}_{B} \in \mathbb{R}^{3}$ are the absolute linear and angular velocities of the base, $\mathbf{q}_{j},\dot{\mathbf{q}}_{j} \in \mathbb{R}^{12}$ are the angular positions and velocities of the joints. The robot is controlled using joint torques $\boldsymbol{\tau}_{j} \in \mathbb{R}^{12}$. Moreover, we assume that we can extract robocentric measurements of \textit{local} terrain elevation via the mapping $\mathbf{M}_{R}: \mathbb{R}^{2} \times \mathbb{R} \rightarrow \mathbb{R}^{32 \times 32}$ with a resolution of $\unit[4]{cm}$. In order to reason precisely about gaits and transitions between contact supports, we define a parameterization thereof that encompasses all necessary information. We thus parameterize a gait as a sequence of so-called \textit{support phases}. Each phase is defined by the tuple
\begin{equation}
\boldsymbol{\Phi} \coloneqq \langle \mathbf{R}_{B},\, \mathbf{r}_{B},\, \mathbf{v}_{B},\, \mathbf{r}_{F},\, \mathbf{c}_{F},\, t_{E},\, t_{S} \rangle \; \in \mathit{\Phi}
\end{equation}
where $\mathbf{c}_{F} \in \{0,1\}^{4}$ is a vector indicating for each of the feet a closed, $1$, or open, $0$, contact w.r.t the terrain, $\mathbf{r}_{F} \in \mathbb{R}^{3 \times 4}$ are the stacked absolute positions of the feet, and $t_{E}, t_{S} \in \mathbb{R}$ are the phase timing variables. For every phase $\boldsymbol{\Phi}_{t}$, $t-t_{E}$ defines the time at which the switch to the current contact configuration occurred, while $t+t_{S}$ the switch to next. Fig.~\ref{fig:Deployment}\,(b) illustrates the aforementioned quantities.

\section{Methodology}
\label{sec:Methodology}

\begin{figure*}
    \vspace{1mm}
    \begin{minipage}[t]{0.74\linewidth}
        \vspace{0mm} 
        \centering
        \includegraphics[width=0.95\linewidth]{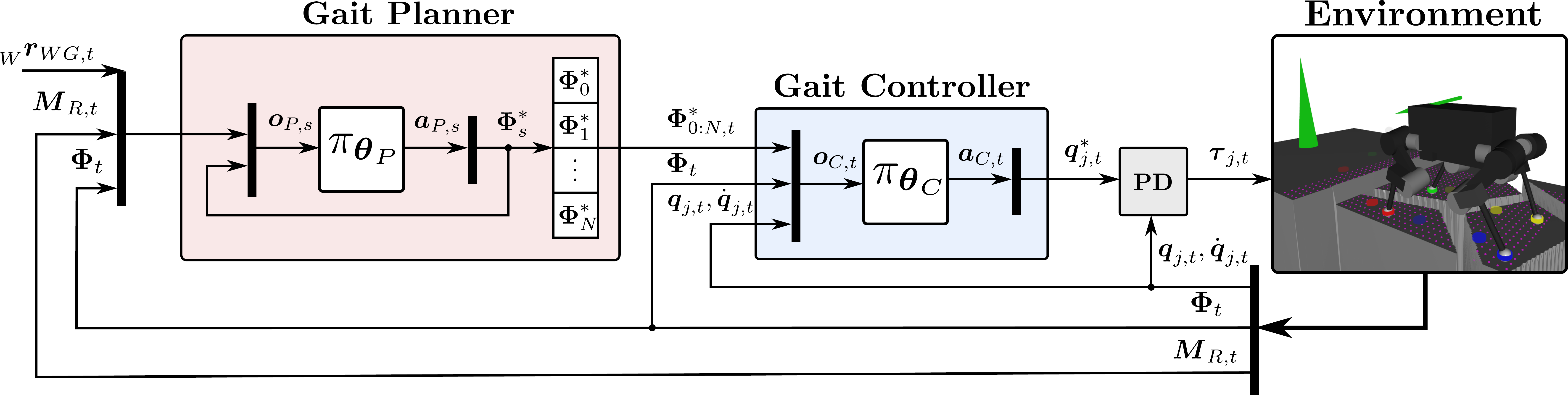}
        \footnotesize{(a)}
    \end{minipage}%
    \hfill
    \begin{minipage}[t]{0.24\linewidth}
        \vspace{0mm} 
        \centering
        \includegraphics[width=0.95\linewidth]{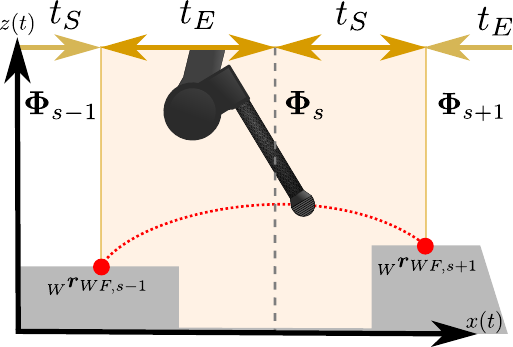}
        \footnotesize{(b)}
    \end{minipage}%
    \caption{\textbf{(a)} Overview of the proposed control structure used at deployment time.
    \textbf{(b)} Phases within a sequence are indexed using $s$, and every index corresponds to a point in time centered around a window defined by the durations $t_{E}$ and $t_{S}$. The center of the window is defined by the motion of the base as captured by the phase $\boldsymbol{\Phi}_{s}$. $t_{S}$ defines the time-to-switch from the current contact support to the next, specified in $\boldsymbol{\Phi}_{s+1}$, and $t_{E}$ defines the time elapsed since the switch from the previous contact support, specified in $\boldsymbol{\Phi}_{s-1}$, to the current.}
    \label{fig:Deployment}
\end{figure*}

We propose a two-level hierarchy comprising a high-level Gait Planner (GP) and a low-level Gait Controller (GC) operating at different time-scales, inspired by~\cite{Peng:2017:DDL:3072959.3073602}. The GP, evaluated at roughly 2Hz, serves as a local terrain-aware planner, and uses both \emph{exteroceptive} and \emph{proprioceptive} measurements to generate a finite sequence of support phases, i.e. a \emph{phase plan}. The GC, evaluated at 100Hz, serves as a hybrid motion planner and controller, and uses \emph{only proprioceptive} sensing in combination with the aforementioned phase plan to output joint position references. Finally, a joint-space PD controller (with zero target joint velocity) uses these joint position references to compute joint torques at 400Hz. The highest-level command to the system is provided as the desired walking direction in the form of the deviation of base attitude w.r.t the goal. Fig.~\ref{fig:Deployment}(a) provides an overview of the system.

\subsection{Gait Planning}
\label{subsec:GaitPlanningMethod}
The GP operates by sequentially querying the planning policy $\pi_{\boldsymbol{\theta}_{P}}$ to generate the aforementioned phase plan. We thus formulate an MDP in order to train $\pi_{\boldsymbol{\theta}_{P}}$ using DRL, and our objective is to ensure that the resulting policy learns to respect the kinodynamic properties and limits of the robot, as well as contact constraints, when proposing phase transitions. Moreover, we aim to avoid direct interaction with the physics of the system, and instead craft the transition dynamics of the MDP by employing a \textit{transition feasibility} criterion realized as a Linear Program (LP) using the framework defined in~\cite{fernbach2019ccroc}. Lastly, we avoid explicitly modeling or qualifying the terrain, as done in~\cite{fankhauser2018robust,tonneau2018efficient}, and instead directly use measurements of local terrain elevation. The resulting MDP, allows us to train $\pi_{\boldsymbol{\theta}_{P}}$ to infer a distribution over phase transitions, which are not only feasible but also maximize locomotion performance (refer to the discussion in Sec.~\ref{sec:Discussion}).

\paragraph*{Support Phase Transition Feasibility} Transition feasibility amounts to evaluating if a feasible motion exists between a pair of support phases $\boldsymbol{\Phi}_{t}, \boldsymbol{\Phi}_{t+1}^{*}$, where the former is assumed while the latter is a candidate successor. As previously mentioned, we employ the general framework defined in~\cite{fernbach2019ccroc} to design a convex LP using the Convex Resolution Of Centroidal dynamics trajectories (CROC) formulation. We use CROC to derive a set of linear equality and inequality constraints, forming a convex polytope, using the following terms:
\begin{enumerate}
    \item A Centroidal Dynamics model of the system.
    \item The contact force unilateral and friction constraints.
    \item Assume angular momentum rate of zero.
    \item Parameterization of CoM motion as a Bezier curve.
    \item Restrict motion of the feet w.r.t. the base.
    \item Restrict contact forces in magnitude and direction.
\end{enumerate}
The specific variant of CROC we employ uses a trivial cost, a time-discretization of the CoM trajectory, and incorporates the parameterization of the contact forces into the decision variables of the optimization. We defer the reader to~\cite{fernbach2019ccroc} for further details regarding CROC. The resulting formulation allows us to realize the transition feasibility mapping $\mathit{F}_{croc}: \mathit{\Phi} \times \mathit{\Phi} \rightarrow \left\{0,1\right\}$. Therefore, we evaluate the LP for pairs of phases $\boldsymbol{\Phi}_{t},\boldsymbol{\Phi}_{t+1}^{*}$, to determine if the corresponding phase transition is feasible, 1, or not, 0.

\paragraph*{MDP Specification}
Modeling locomotion as discrete sequences of support phases $\boldsymbol{\Phi}$, as defined in Sec.~\ref{subsec:SystemModel}, allows us to formulate phase transitions that exhibit the Markov property~\cite{Sutton:2018}, and thus lends itself to modeling the overall problem of gait planning as an MDP. Specifically, we define the state of the MDP as the tuple
$\boldsymbol{s}_{P} \coloneqq \langle \boldsymbol{\Phi}, \boldsymbol{r}_{G} \rangle$,
where $\boldsymbol{r}_{G}$ is the current absolute position of the goal in the world, while observations and actions are defined as the tuples
$\boldsymbol{o}_{P} \coloneqq \langle \boldsymbol{o}_{R},\boldsymbol{o}_{v},\boldsymbol{o}_{F},\boldsymbol{o}_{c},\boldsymbol{o}_{M}\rangle$
and
$\boldsymbol{a}_{P} \coloneqq \langle \boldsymbol{a}_{R},\boldsymbol{a}_{B},\boldsymbol{a}_{v},\boldsymbol{a}_{F},\boldsymbol{a}_{c},\boldsymbol{a}_{t} \rangle$,
respectively. Observations, consist of terms pertaining to the current state of the robot and the coincident terrain in the form of: the attitude w.r.t the goal $\boldsymbol{o}_{R}\in \mathbb{R}$, the CoM velocity $\boldsymbol{o}_{v}\in \mathbb{R}^2$, the feet positions $\boldsymbol{o}_{F}\in \mathbb{R}^8$, the feet contact states $\boldsymbol{o}_{c}\in \mathbb{R}^4$ and the local height-map $\boldsymbol{o}_{M}\in \mathbb{R}^{32 \times 32}$. Conversely, actions, contain terms pertaining to changes to the current phase $\boldsymbol{\Phi}$, in the form of the CoM rotation $\boldsymbol{a}_{R}\in \mathbb{R}_{clip}$, CoM translation $\boldsymbol{a}_{B}\in \mathbb{R}_{clip}^2$, CoM velocity $\boldsymbol{a}_{v}\in \mathbb{R}_{clip}^2$, feet positions $\boldsymbol{a}_{F}\in \mathbb{R}_{clip}^8$, feet contact states $\boldsymbol{a}_{c}\in \mathbb{R}_{clip}^3$ and the phase timings $\boldsymbol{a}_{t}\in \mathbb{R}_{clip}^2$. All action terms are scaled, offset and clipped to lie in $\mathbb{R}_{clip} \coloneqq \left[-1;1\right]$. The exact definitions for actions and observations are provided in Tab.~\ref{table:GpObservationsActions}.
\begin{table}[t]
\vspace{6mm}
    \centering
    \caption{Transformations used to form $\boldsymbol{o}_{P}$ from phases and phase transitions from $\boldsymbol{a}_{P}$: the matrix ${\boldsymbol{R}_z}(\alpha)$ defines a rotation about the world's z-axis by angle $\alpha$, and $f_{dec}: \mathbb{R}_{clip}^3 \rightarrow \left\{0,1\right\}^4$ decodes a 3-digit binary encoding into a vector of contact states.}
    \begin{tabular}{c|c}
        \hline
        \textbf{GP Observation} & \textbf{GP Action} \\ 
        \hline
         \parbox{3.5cm}{
         \begin{align*}
            &\boldsymbol{o}_{v}=\prescript{}{B}{\boldsymbol{v}}_{{B}_{x,y}}
            \\
            &\boldsymbol{o}_{F}=\prescript{}{B}{\boldsymbol{r}}_{{BF}_{x,y}} - \prescript{}{B}{\boldsymbol{r}}_{{BN}_{x,y}}
            \\
            &\boldsymbol{o}_{c}= 2 \, \boldsymbol{c}_{F} - \mathbf{1}_{4\times1}\\
            &\boldsymbol{o}_{M}= \boldsymbol{M}_{R} \\
            &\boldsymbol{o}_{R}= -atan2\Big(\frac{\prescript{}{B}{\boldsymbol{r}}_{BG,y}}{\prescript{}{B}{\boldsymbol{r}}_{BG,x}}\Big)
        \end{align*}
         } &
         \parbox{3.5cm}{
         \begin{align*}
            \boldsymbol{R}_{B}^{*} &= {\boldsymbol{R}_z}\big(\frac{\pi}{8}\boldsymbol{a}_{R}\big) \boldsymbol{R}_{B}\\
            \prescript{}{B}{\boldsymbol{r}}_{{B}_{x,y}}^{*}&=\prescript{}{B}{\boldsymbol{r}}_{{B}_{x,y}} + 0.3 \, \boldsymbol{a}_{B}\\
            \prescript{}{B}{\boldsymbol{v}}_{{B}_{x,y}}^{*}&=\boldsymbol{a}_{v},\;
            \boldsymbol{c}_{F}^{*} = f_{dec}(\boldsymbol{a}_{c})\\
            \prescript{}{B}{\boldsymbol{r}}_{{BF}_{x,y}}^{*}&= \prescript{}{B}{\boldsymbol{r}}_{{BN}_{x,y}} + 0.3 \, \boldsymbol{a}_{F} \\
            \begin{bmatrix}
                t_{E}^{*}, t_{S}
         \end{bmatrix}^{T} &=
         \mathbf{1}_{2\times1} + 0.9 \, \boldsymbol{a}_{t}
        \end{align*}}\\
        \hline
    \end{tabular}
    \label{table:GpObservationsActions}
\vspace{-4mm}
\end{table}
\paragraph*{Transition Dynamics}
We define state transition dynamics for this MDP employing a formalism defining so-called \textit{termination condition} functions $\mathit{T}(\boldsymbol{s}_{P,t}, \boldsymbol{a}_{P,t}, \boldsymbol{s}_{P,t+1})$, which determine if an episode terminates. By formulating an episode termination as a transition into an absorbing terminal state, we can say that, an episode under this MDP, terminates whenever $\boldsymbol{s}_{P,s+t} = \boldsymbol{s}_{P,s},\, \forall t>0$.
In this MDP, in particular, we employ the following termination conditions:
\begin{enumerate}
    \item $\mathit{T}_{footholds}$: Checks for obstacles or gaps within the vicinity of each foothold using an fixed eight-point grid surrounding each foot.
    \item $\mathit{T}_{base}$: Checks for collisions between base and terrain.
    \item $\mathit{T}_{feasibility}$: Evaluates $\mathit{F}_{croc}(\boldsymbol{\Phi}_{t}, \boldsymbol{\Phi}_{t+1}^{*})$.
\end{enumerate}
Thus, each step of this MDP proceeds as follows:
(1) Given a state $\boldsymbol{s}_{P,t}$, the MDP computes the corresponding observation $\boldsymbol{o}_{P,t}$, which is constructed according to the transformations in Tab.~\ref{table:GpObservationsActions}, and is passed to the agent to select an appropriate action according to $\pi_{\boldsymbol{\theta}_{P}}$.
(2) The selected action $\boldsymbol{a}_{P,t}$, is used to compute the candidate phase $\boldsymbol{\Phi}_{t+1}^{*}$, again using the set of transformations defined in Tab.~\ref{table:GpObservationsActions}.
(3) The aforementioned terminations conditions are used to assert if the phase transition is feasible. This formulation therefore allows the agent to propose the phase transition directly, while the MDP only checks if it is feasible and otherwise terminate the episode.

In addition, we outline certain considerations regarding the computation of $\boldsymbol{\Phi}_{t+1}^{*}$. First, if a foot would be in contact for both phases $\boldsymbol{\Phi}_{t}$ and $\boldsymbol{\Phi}_{t+1}^{*}$, the new foothold is reset to that of $\boldsymbol{\Phi}_{t}$ since stance feet cannot change positions. Second, the z-coordinates of the feet and CoM positions are adjusted according to the height of the terrain at their respective locations, and for the CoM in particular, the height is set to a constant $h_{com}$ above the lowest foothold. Lastly, to evaluate the kinematic constraints of CROC, we need to infer the intermediate orientations. To this end, we use a first-order approximation of the angular velocity of the base by linearly interpolating between the starting and next base attitudes of each transition. Doing so also renders zero angular momentum between consecutive support phases, which is inline with assumptions made by CROC. One can thus envision an extension to CROC that includes angular momentum, which we intend to explore in future work.

\paragraph*{Reward Function}
We design a reward function which drives the agent to learn behaviors for reaching the goal position, facing the goal as much as possible, minimizing kinematic effort during phase transitions and inhibiting long stance phases. The final reward function is specified as the combination of multiplicative and additive terms
\begin{equation}
    r_{P}(\boldsymbol{s}_{P,t}, \boldsymbol{s}_{P,t}, \boldsymbol{s}_{P,t+1}) \coloneqq r_{p} \cdot r_{h}^{2} \cdot r_{k} - r_{c}
\end{equation}
where $r_{p}$ rewards the agent for bringing the average foothold position closer towards the goal and penalizes moving it away, $r_{h}$ penalizes the robot for not facing the goal, $r_{k}$ penalizes for moving the feet away from the nominal footholds $\prescript{}{B}{\boldsymbol{r}}_{NF}$ located beneath the shoulders and $r_{c}$ penalizes for not lifting a foot over multiple steps, therefore promoting exploration and prevents the policy from getting stuck in the local optimum of remaining in a constant stance.

Furthermore, we would like to emphasize certain features of the multiplicative term in the above reward function. Specifically, this term results in a penalty that is small when $r_{p}$ is small, i.e., beginning of training, and large when $r_{p}$ is large, i.e., towards the end of training, thus resulting in a form of automatic scaling of the overall multiplicative term. We found that using these multiplicative rewards results in beneficial gradients throughout all iterations of training, as their values are ensured never to be too large as to hinder exploration, and never too small as to have negligible effect. Furthermore, as $r_{p}$ is computed using the average position of footholds and not of the base, the agent is required to walk in order to maximize reward, as opposed to just merely leaning. The latter aspect is important, since leaning forward also inhibits the motion of the front legs, therefore making it much harder to walk. Tab.~\ref{table:RewardTable} details the aforementioned reward terms.

\begin{figure}[t]
    \centering
    \includegraphics[width=\linewidth]{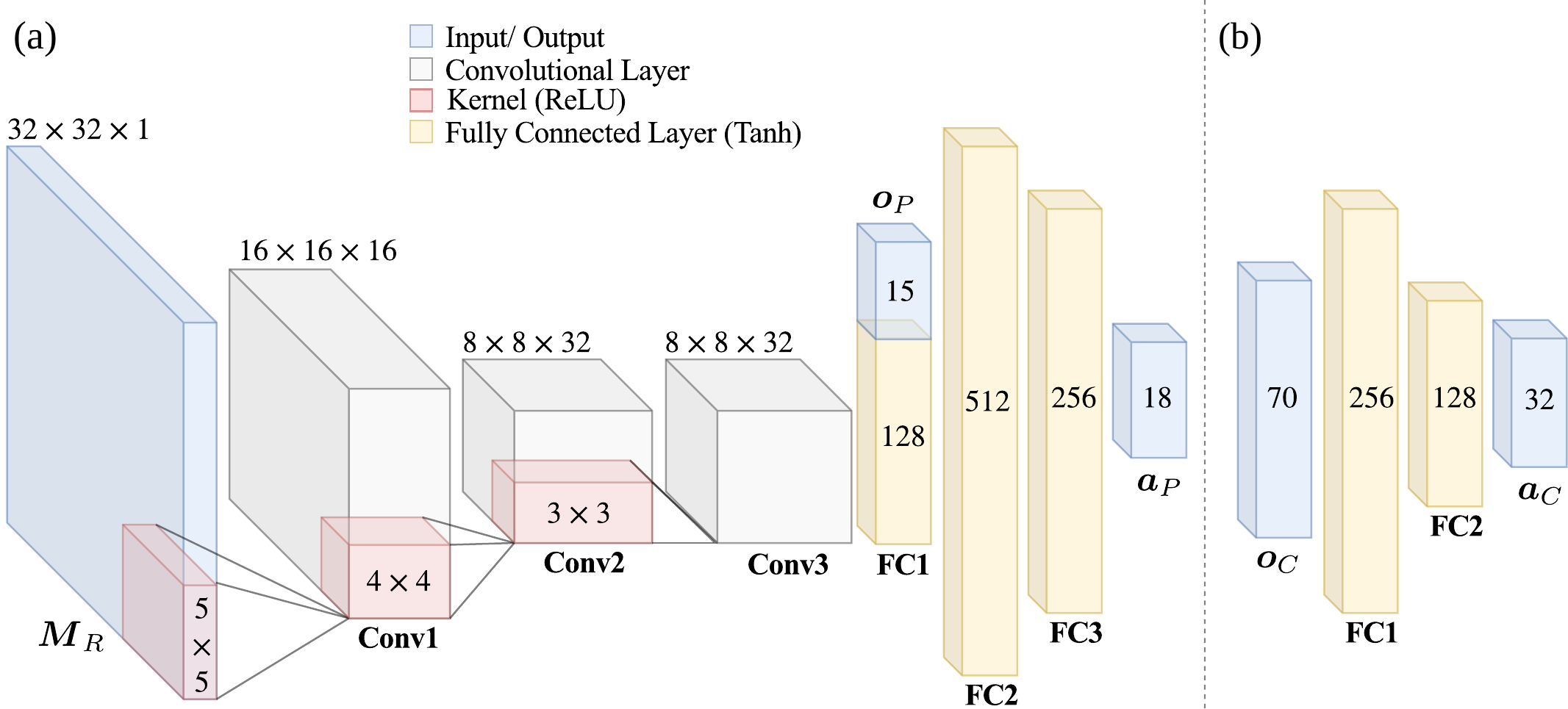}
    \caption{The neural-network models used for the latent parameters of policy distributions of the (a) GP  and  (b) GC, respectively.}
    \label{fig:Networks}
\end{figure}

\paragraph*{Policy Definition}
We parameterize the GP's policy as a Gaussian distribution with a diagonal covariance matrix $\pi_{\boldsymbol{\theta}_{P}}(\boldsymbol{a}|\boldsymbol{o}_{P,t}) \coloneqq \mathcal{N}(\boldsymbol{a} | \boldsymbol{\mu}_{\boldsymbol{\theta}_{P}}(\boldsymbol{o}_{P,t}), \boldsymbol{\sigma}_{\boldsymbol{\theta}_{P}})$. The mean $\boldsymbol{\mu}_{\boldsymbol{\theta}_{P}}(\boldsymbol{o}_{P,t})$ is output by a NN which inputs both exteroceptive and proprioceptive measurements into a series of NN layers, similar to those proposed in \cite{Peng:2017:DDL:3072959.3073602}. First, $\boldsymbol{M}_{R}$ is input into three CNN layers, the output of which is subsequently input into one more fully-connected layer. The resulting latent output from the height-map is concatenated with the raw proprioceptive measurements, then fed into two more fully-connected layers with ReLU and tanh nonlinearities, and finally passed through a  linear output layer. The standard-deviation parameters $\boldsymbol{\sigma}_{\boldsymbol{\theta}_{P}}$ are realized by an additional layer that is independent of the observations and is used to drive exploration during training. Fig.~\ref{fig:Networks}(a) provides a graphical depiction of the NN model. Due to the inclusion of high-dimensional height-map data in the observations $\boldsymbol{o}_{P}$ as well as the relatively large dimensionality of the actions $\boldsymbol{a}_{P}$, we trained $\pi_{\boldsymbol{\theta}_{P}}$ with a variant of Proximal Policy Optimization (PPO) using clipped loss and a Generalized Advantage Estimation (GAE) critic~\cite{schulman2017proximal}.

\subsection{Gait Control}
\label{subsec:GaitControlMethod}
The GC is responsible for executing the support phase sequence provided by the GP while maintaining balance at all times. It operates by tracking a series of footholds and base positions extracted from the support phase sequence generated by the GP. In order to learn this behavior, we define an MDP with transition dynamics which incorporates the physics of the system and specify an appropriate parameterization for $\pi_{\boldsymbol{\theta}_{C}}$. Training a GC agent in such an MDP requires only that a target phase sequence be provided, and does not assume that a GP is available a priori. In fact, the target phase sequence can be provided arbitrarily as long as the target footholds are feasible. However, in this work, we elected to utilize a GP for this purpose solely as a matter of convenience so to avoid the use of additional elements.

\paragraph*{Target Foothold Extraction} Assuming the GP is queried at some time $t$, we denote the resulting phase sequence as $\boldsymbol{\Phi}_{0:N,t}^{*}$, where $\boldsymbol{\Phi}_{0,t}^{*}$ is the initial phase as measured by the GP before generating the sequence of length $N$. This amounts to \textit{rolling-out}\footnote{The small range and dimensions selected for the elevation map, in conjunction with the limitation on maximum step length assumed by the planner, allows us to extract multiple successive samples of $\mathbf{M}_{R}$ from within the effective field-of-view afforded by exteroceptive sensing.} the planning policy by recursively evaluating $\pi_{\boldsymbol{\theta}_{P}}$ using its own output.

\paragraph*{MDP Definition} Given a phase sequence $\boldsymbol{\Phi}_{1:N}^{*}$, the GC proceeds to extract the following target quantities: (a) target position for the base $\mathbf{r}_{B}^{*}$, (b) target feet contact states $\mathbf{c}_{F}^{*}$, and (c) target foothold positions $\mathbf{r}_{F}^{*}$ for all legs. In the case of $\mathbf{r}_{F}^{*}$, targets are set by looking ahead into the phase plan so to ensure that both swing and stance legs have valid references at all times. Thus, the GC computes the foothold tracking errors $_{B}\mathbf{r}_{F,err}$, at 100Hz, while the target footholds are updated at approximately 2Hz. We specify MDP states $\boldsymbol{s}_{C}$, observations $\boldsymbol{o}_{C}$ and actions $\boldsymbol{a}_{C}$ defined as
\begin{align}
        \boldsymbol{s}_{C} \coloneqq &
        \langle
        \mathbf{R}_{B},\,
        \mathbf{r}_{B},\,
        \mathbf{v}_{B},\,
        \boldsymbol{\omega}_{B},\,
        \mathbf{q}_{j},\,
        \dot{\mathbf{q}}_{j},\,
        \mathbf{n}_{F},\,
        \mathbf{c}_{F}
        \rangle
        \notag
        \\
        \boldsymbol{o}_{C} \coloneqq &
        \langle
        _{B}\mathbf{r}_{F,err},\,
        \mathbf{c}_{F}^{*},\,
        _{B}\mathbf{e}_{z}^{W},\,
        z_{BF},\,
        _{B}\mathbf{v}_{B},\,
        \notag
        \\
        \phantom{\boldsymbol{s}_{C} \angle} & \quad
        _{B}\boldsymbol{\omega}_{B},\,
        \mathbf{c}_{F},\,
        \mathbf{q}_{j},\,
        \dot{\mathbf{q}}_{j},\,
        \mathbf{q}_{j}^{*},\,
        \eta
        \rangle,
        \quad
        \boldsymbol{a}_{C} \coloneqq \langle \mathbf{q}_{j}^{*} \rangle
    \label{eq:MdpGc}
\end{align}
where $_{B}\mathbf{e}_{z}^{W}$ is the gravity-aligned z-axis of frame $W$ expressed in coordinates of frame $B$, $z_{BF}$ is the distance between the lowest stance foot and the base along the z-axis of $W$, $\mathbf{q}_{j}^{*}$ is the vector of previous target joint positions, and $\eta \in [0,1]$ is a phase variable indicating the normalized time within a support phase. The transition dynamics of this MDP includes the generation of the phase plan using the GP, the physics of the system and the joint-space PD controller. As the PD controller is evaluated at 400Hz and the GC at 100Hz, we apply a zero-order hold of the joint positions output by the policy when computing joint torques commands. For this MDP, we also define the following termination conditions:
\begin{enumerate}
    \item $\mathit{T}_{attitude}$: Angle between $\boldsymbol{e}_{z}^{B}$ and $\boldsymbol{e}_{z}^{W}$ exceeds $60^{\circ}$.
    \item $\mathit{T}_{contact}$: Base collides with the terrain.
\end{enumerate}
We have found that these two simple, yet effective, termination conditions are those principally responsible for the balancing and recovery behaviors learned during training, and thus their importance must not be understated. Moreover, we designed a reward function that emphasizes tracking of target foothold positions and contact states, but also contains terms that inhibit extraneous and aggressive motions during locomotion. The resulting reward function is defined as
\begin{align}
        r_{C}(\mathbf{s}_{C,t}, \mathbf{a}_{C,t}) \, \coloneqq
        r_{e} + r_{tc} + r_{sw} + r_{sl}
        + r_{t} + r_{v} + r_{a}
\end{align}
where $r_{e}$ and $r_{tc}$ are the task-specific rewards penalizing deviations from the target foothold positions and contact states, $r_{sl}$ penalizes foot-slip for grounded feet, $r_{sw}$ penalizes large velocities for swing legs, $r_{t}$ penalizes joint torques, $r_{v}$ penalizes vertical linear and roll-pitch angular velocities of the base and $r_{a}$ penalizes large angles between the unit vectors $\boldsymbol{e}_{z}^{B}$ and $\boldsymbol{e}_{z}^{W}$ of the base and world frame respectively.

\paragraph*{Policy Definition} The GC's policy, like that of the GP, is also parameterized as a Gaussian distribution with diagonal covariance matrix $\pi_{\boldsymbol{\theta}_{C}}(\boldsymbol{a}|\boldsymbol{o}_{C,t}) \coloneqq \mathcal{N}(\boldsymbol{a} | \boldsymbol{\mu}_{\boldsymbol{\theta}_{C}}(\boldsymbol{o}_{C,t}), \boldsymbol{\sigma}_{\boldsymbol{\theta}_{C}})$. While the mean $\boldsymbol{\mu}_{\boldsymbol{\theta}_{C}}(\boldsymbol{o}_{C,t})$ is output by a simple NN with two fully-connected layers using $tanh$ non-linearities, shown in Fig.~\ref{fig:Networks} (b), the standard deviation coefficients $\boldsymbol{\sigma}_{\boldsymbol{\theta}_{C}}$ are, just as in the case of the GP, output by an additional layer of parameters which is independent of $\boldsymbol{o}_{C,t}$. Due to the relatively small dimensionality of the MDP and $\pi_{\boldsymbol{\theta}_{C}}$, we train the latter using Trust-Region Policy Optimization (TRPO) also employing a GAE critic~\cite{schulman2015high}.

\begin{table*}[t]
    \vspace{6mm}
    \centering
    \caption{The GP and GC reward terms. The subscript $k$ indexes each foot, and $n_{contact,k}$ counts the current number of consecutive state updates for which foot $k$ has remained grounded. $b_{k}$ is a binary variable specifying whether a foot is within a cylinder of radius $d=\unit[5]{cm}$ of target foothold. An over-lined letter describes the conjugate of the binary variable, e.g. $\Bar{b}_k$. The weighting factors are: $w_{p}=25$, $w_{k}=80$, $w_{c}=0.01$, $w_{tc}=0.1$, $w_{sw}=0.01$, $w_{t}=0.001$, $w_{v}=0.5$, $w_{e}=2$, $w_{sl}=0.02$, $w_{a}=0.2$.}
     \begin{tabular}{c|c}
        \hline
        \textbf{Gait Planner Rewards} & \textbf{Gait Controller Rewards} \\ 
        \hline
         \parbox{2.0cm}{
         {\scriptsize
         \vspace{-2mm}
         \begin{align*}
            r_{p} &= w_{p}\norm{\boldsymbol{r}_{G}-\sum_{k=1}^4 \boldsymbol{r}_{F,k} \boldsymbol{c}_{F,k}}_{2}\!\!-w_{p}\norm{\boldsymbol{r}_{G}-\sum_{k=1}^4 \boldsymbol{r}_{F,k}^{*} \boldsymbol{c}_{F,k}^{*}}_{2}\\[-2pt]
            r_{k} &= \max\left[0, 1-w_{k}\sum_{k=1}^4 \norm{\prescript{}{B}{\boldsymbol{r}}_{NF,k,x}}_{2}^{3}+ \norm{\prescript{}{B}{\boldsymbol{r}}_{NF,k,y}}_{2}^{3}\right]\\[-4pt]
             r_{h} &= 1-\frac{1}{\pi}\left\vert atan2\left(\frac{\prescript{}{B}{\boldsymbol{r}}_{BG,y}}{\prescript{}{B}{\boldsymbol{r}}_{BG,x}}\right)\right\vert,
             \, r_{c} = w_{c} \sum_{k=1}^4 n_{c,k}
        \end{align*}}
         } &
         \parbox{5.0cm}{
         {\scriptsize
         \vspace{-2mm}
         \begin{align*}
                r_{tc} &= w_{tc}\sum_{k=1}^4 \left[ b_k(c_{F,k}-\Bar{c}_{F,k})+ \Bar{b}_k(c_{F,k}(c_{F,k}^* - \Bar{c}_{F,k}^*)+\Bar{c}_{F,k}(\Bar{c}_{F,k}^* - c_{F,k}^*))\right]\\[-8pt]
                r_{sw} &= - w_{sw}\sum_{k=1}^4 \left[ \Bar{c}_{F,k}\norm{\boldsymbol{v}_{F,k}}_{2}^{2} -\Bar{b}_k \Bar{c}_{F,k}^*\min(d, \boldsymbol{r}_{F,k,z} - \boldsymbol{r}_{F,k,z}^*)  \right]\\[-8pt]
                r_{t} &= - w_{t} \norm{\boldsymbol{\tau}}_{2}^{2},\, r_{a} = -w_{a} acos((\boldsymbol{e}_{z}^{B})^T\!\boldsymbol{e}_{z}^{W}),\, r_{sl} =- w_{sl} \sum_{k=1}^4 c_{F,k}\norm{\boldsymbol{v}_{F,k,xy}}\\[-8pt]
                r_{e} &= -w_{e}\sum_{k=1}^4\sqrt{\norm{\boldsymbol{r}_{F,k}^*-\boldsymbol{r}_{F,k}}_{2}}, \, r_{v} = -  w_{v}\lvert\prescript{}{B}{\boldsymbol{v}}_{B,z}\rvert^{2} - \norm{\prescript{}{B}{\boldsymbol{\omega}}_{B,xy}}_{2}^{2}
        \end{align*}}}\\
        \hline
    \end{tabular}
    \label{table:RewardTable}
    \vspace{-5mm}
\end{table*}

\section{Results}
\label{sec:Results}

\subsection{Experimental Setup}
\label{subsec:ExperimentalSetup}
In order to evaluate our approach, we crafted a suit of terrain scenarios for training and testing the GC and GP agents, as depicted in Fig.~\ref{fig:TerrainSuite}. The first and most basic scenario consists of an infinite flat plane we refer to as \textit{Flat-World}, which we use to establish a baseline for performance and behavior. Secondly, the \textit{Random-Stairs} terrain presents a $\unit[20\times20]{m^2}$ square area consisting of $\unit[1\times1]{m^2}$ flat regions of randomly selected elevation. The elevation changes were generated in a way that results in an effective inclination diagonally across the map. The third terrain scenario is that which we call \textit{Temple-Ascent}, and is a composite terrain consisting of gaps, stepping stones, stairs as well as flat regions. We realized the MDP environment for the GP using an own implementation of CROC in C++, while for the MDP environment of the GC we used the  RaiSim~\cite{raisim} multi-body physics engine. All DRL algorithms were implemented using the \textit{TensorFlow}\footnote{\href{https://github.com/leggedrobotics/tensorflow-cpp}{\url{https://github.com/leggedrobotics/tensorflow-cpp}}} C/C++ API\footnote{For the GP, we used a PC with 2x Intel Xeon E5-2680v4 (@2.4GHz) CPUs, 128GB of RAM, and an Nvidia GTX Titan (Pascal), and for the GC a PC with a single Intel Core i7-8700K (@3.7GHz) CPU, 64GB of RAM and an  Nvidia GTX 2080 Ti GPU}.

\subsection{Gait Planner}
\label{subsec:GpResults}

\paragraph*{Training Setup} Training of GP policies in the terrain suite consists of a set of episodes where the robot's objective is to reach a a goal position from a sufficiently distant starting location. Both starting and goal positions are selected randomly at the start of each episode. However, this procedure differs depending on the features of the terrain, as we must avoid invalid starting positions and unreachable goal positions, which, would negatively impact the resulting policies during training. Once valid starting and goal positions have been sampled, the robot's initial attitude and footholds are also sampled uniformly from within respective bounds.

We thus trained two separate GP policies for \textit{Random-Stairs} and \textit{Temple-Ascent} respectively, using PPO with only $14$ parallel workers running on the respective desktop computer over $200k$ iterations, which amounts to a total of two billion samples per run. Hyper-parameter values are listed in Tab.~\ref{table:HyperParams}. We did not need to train a separate GP in \textit{Flat-World}, and instead used that trained in \textit{Temple-Ascent} for the respective performance evaluations.

\paragraph*{Performance Metrics} In order to assess the performance of GP policies, we define the Episodic Success Rate (ESR), which measures the number of successfully reached goal positions over a finite number of episodes. Essentially, we execute a sufficiently large number of episodes where the robot tracks a reference goal position in the world and assert if the robot has reached within a $\unit[0.5]{m}$ vicinity of the goal and within a maximum permissible episode duration.

\paragraph*{Training Results} GP training required approximately $82$ hours in each terrain scenario. Throughout our experiments we found that the randomization scheme mentioned above and used for realizing the initial state distribution of the MDP was crucial for successfully learning to traverse all parts of the terrains. This demonstrates that if the agent does not observe all aspects of the terrain from the very beginning of training, it is often unable to generalize to unseen cases at test time.

Furthermore, we observed that, as the centroidal dynamics model employed by CROC is relatively conservative, it tends to limit the set of transitions that the policy can generate. This conservativeness is furthered by the fact that in this work, we limit the possible contact states that the GP's policy can output to only those with three and four active contacts. Such a restriction was helpful for reducing the complexity of the problem, and we intend to extend to the general case of two and single contact configurations in future work.

We tested the GP policies in their respective terrain scenarios and evaluated their performance using the ESR metric. In all cases, we have observed that fully trained policies can generate valid support phase sequences which lead the robot to the goal with at an average ESR nearing $\unit[100.0]{\%}$. The performance of GP policies trained and tested in the terrain suite are presented in Tab.~\ref{table:Performance}, where they have been deployed together with respective GC policies.

Another important observation regarding the output of the GP has to do with the types of gaits it manifests. In the case of \textit{Flat-World} as well as in the flat regions of \textit{Temple-Ascent}, we observe that the GP tends to output mostly cyclic support phases. This indicates that the agent learns to generate cyclic gaits even though no aspect of the MDP ever directed it to do so. In certain cases, however,  such as the \textit{Stepping-Stones} and \textit{Gaps} bridges as well as when performing sharp point-turns, the GP outputs acyclic support phases.

\paragraph*{Sample Complexity} One key contribution of this work is the significant reduction in sample complexity afforded by the use of transition feasibility instead of physical simulation to formulate the GP's MDP. Using the transition feasibility check, we can evaluate the MDP's transition dynamics at several thousands of steps-per-second, where each step corresponds to potentially several seconds of simulation time. Conversely, using a physics simulator typically requires several hundred or even thousands of steps to evaluate just one second of simulation time. Specifically, during training, we executed episodes with a maximum length of $50$ steps with each corresponding to an average duration of approximately $\unit[2.6]{s}$, which amounts to $\unit[130]{s}$ of simulation time. However, the physics simulator using a time-step of $\unit[2.5]{ms}$ would require $24k$ steps to simulate the duration above. As the throughput of the transition feasibility LP and the physics simulator, for our formulation, is $\unit[1k]{Hz}$ and $\unit[60k]{Hz}$ respectively, we can estimate an $18$-fold effective reduction in sample-complexity.

\subsection{Gait Controller}
\label{subsec:GcResults}
\begin{table*}[ht!]
    \vspace{6mm}
    \begin{minipage}[t]{0.6\linewidth}
        \centering
        \caption{Performance of the GC on the different terrain scenarios in \textit{Temple-Ascent}, and under different kinds of variations to the system. The nominal system is that with which the GC was trained, and all variations are performed only at test time. $m_{B}$ is the mass of the base, while $l_{shank}$ is the length of the shank links. ESR values are listed as percentages, and all results are presented as empirical means plus-minus the corresponding standard deviations.}
        \begin{tabular}{l c c c c c c}
            \hline
            \textbf{System} & \textbf{Metric} & \textbf{Flat}  & \textbf{Gaps}  & \textbf{Stepping-Stones} & \textbf{Stairs} \\
            \hline
            \multirow{2}{1pt}{Nominal}
            & ESR  & $99.8\%\pm0.2\%$ & $96.4\%\pm2.3\%$ & $96.8\%\pm1.2\%$ & $90.6\%\pm6.8\%$ \\
            & FTS  & $0.985\pm0.000$  & $0.967\pm0.000$  & $0.970\pm0.000$  & $0.751\pm0.000$ \\
            & FTER & $0.016\pm0.000$  & $0.023\pm0.000$  & $0.021\pm0.000$  & $0.049\pm0.000$ \\
            \hline
            \multirow{2}{1pt}{$m_{B}\unit[+25]{\%}$}
            & ESR  & $99.4\%\pm0.8\%$ & $94.6\%\pm6.8\%$ & $98.4\%\pm0.8\%$ & $82.4\%\pm14.3\%$ \\
            & FTS  & $0.916\pm0.000$  & $0.906\pm0.000$  & $0.895\pm0.000$  & $0.605\pm0.000$ \\
            & FTER & $0.028\pm0.000$  & $7.332\pm266.3$  & $0.032\pm0.000$  & $0.060\pm0.000$ \\
            \hline
            \multirow{2}{1pt}{$l_{shank}\unit[+10]{\%}$}
            & ESR  & $99.0\%\pm1.5\%$ & $95.2\%\pm8.7\%$ & $97.6\%\pm0.3\%$ & $76.4\%\pm8.8\%$ \\
            & FTS  & $0.968\pm0.000$  & $0.952\pm0.000$  & $0.975\pm0.000$  & $0.618\pm0.000$ \\
            & FTER & $0.020\pm0.000$  & $0.025\pm0.000$  & $0.020\pm0.000$  & $0.069\pm0.000$ \\
            \hline
            \multirow{2}{1pt}{$l_{shank}\unit[-10]{\%}$}
            & ESR  & $100.0\%\pm0.0\%$ & $97.8\%\pm3.2\%$ & $97.4\%\pm0.8\%$ & $89.6\%\pm16.8\%$ \\
            & FTS  & $0.990\pm0.000$  & $0.965\pm0.000$  & $0.971\pm0.000$  & $0.541\pm0.000$ \\
            & FTER & $0.017\pm0.000$  & $0.022\pm0.000$  & $0.021\pm0.000$  & $0.058\pm0.000$ \\
            \hline
        \end{tabular}
        \label{table:Performance}
    \end{minipage}%
    \hfill
    \begin{minipage}[t]{0.37\linewidth}
        \centering
        \caption{Policy optimization algorithm hyper-parameters for the GC using TRPO and the GP using PPO (see \cite{schulman2015trust,schulman2017proximal} for details).}
        \vspace{1.4mm}
        \begin{tabular}{l|c|c|c}
            \hline
            \textbf{Parameter} & \textbf{Symbol} & \textbf{TRPO} & \textbf{PPO} \\ 
            \hline
            Batch Size          & $N_{B}$           & 24k       & 200k \\
            Mini-Batches        & $N_{MB}$          & -         & 5 \\
            Max. Episode Length & $T_{max}$         & $3000$    & $50$ \\
            Discount Factor     & $\gamma$          & 0.995     & 0.99 \\
            Trace Decay         & $\lambda$         & 0.99      & 0.97 \\
            Terminal Reward     & $r_{T}$           & $-5.0$    & $-1.0$\\
            KL Constraint       & $\delta$          & 0.01      & - \\
            Clip                & $\epsilon$        & -         & 0.2 \\
            Entropy Weight      & $\beta$           & $0.001$   & $0.004$ \\
            Initial Variance    & $\sigma_{0}^{2}$  & $0.4$     & $1.0$ \\
            Adam Epochs         & $n_{epoch}$       & -         & $3$\\
            Adam Learning-Rate  & $\alpha_{Adam}$   & -         & $0.0002$\\
            Gradient Clipping   & $g_{max}$         & -         & $1.0$\\
            CG Damping          & $\beta_{CG}$      & 0.1       & - \\
            CG Steps            & $n_{CG}$          & 40        & - \\
            \hline
        \end{tabular}
        \label{table:HyperParams}
    \end{minipage}%
    \vspace{-4mm}
\end{table*}
\paragraph*{Training Setup} Training a GC agent involves collecting MDP transitions over a rich set of target footholds. In order to achieve such a distribution of training samples, we ensure that both the initial state distribution of the MDP as well as the target footholds generated by the GP are appropriately and sufficiently randomized. Initial states are generated by first uniformly sampling \textit{initial} and \textit{goal} positions of the base from within the bounds of the world. We then orientate the base by sampling uniformly from attitudes centered on the current orientation facing the goal, and bounded by the vector of Euler angles $\begin{bmatrix} 0.1, 0.1, \pi/4 \end{bmatrix}$. Moreover, we randomize the initial feet positions by uniformly sampling $xy$ coordinates from a $\unit[0.1\times0.1]{m^2}$ box defined in the base frame $B$ and centered around nominal values that would place the feet below the shoulders. Furthermore, to randomize the target footholds, we perform a randomized fixed rollout of the GP up to the maximum permissible length of an episode. Essentially, we rollout the GP however many times necessary such that the resulting phase sequence meets or exceeds the duration time of an episode, and randomize the target footholds at each step by adding a bias uniformly sampled from $\begin{bmatrix}-0.1, 1.0\end{bmatrix}$ in the $xy$ plane while ensuring that the $z$ coordinates are fixed to the terrain.

With the aforementioned sampling scheme, we trained a GC agent using TRPO in \textit{Flat-World} using only $24$ parallel workers for total of $20k$ iterations. Moreover, as part of ongoing work to extend our method to full 3D foothold tracking, we present preliminary results for GC policies for stair-climbing by first pre-training in \textit{Random-Stairs} then also on the stairs section of \textit{Temple-Ascent}. Tab.~\ref{table:HyperParams} presents the hyper-parameters most pertinent to the training of GC, for all of the cases mentioned above. We want to emphasize that in all cases, the same hyper-parameters were used, as we only adapted the initial state distribution accordingly for each terrain. TRPO was employed using mostly the default hyper-parameters specified in \cite{schulman2015high,schulman2015trust}.

\paragraph*{Performance Metrics} We define two metrics for quantifying the performance of GC policies at test-time. First we define the \textit{Foothold Tracking Error Rate} (FTER)
\begin{equation}
    FTER \coloneqq  \frac{1}{T} \sum_{t=0}^{T} \frac{1}{\sum_{k=1}^{4}c_{F,k}^{*}} \sum_{k=1}^{4} c_{F,k}^{*} \lVert \boldsymbol{r}_{F,k}^{*} - \boldsymbol{r}_{F,k} \rVert_{2}
\end{equation}
that measures the mean foothold tracking error throughout an individual episode of length $T$, and is computed as a function of the desired contact states $\boldsymbol{c}_{F,k}^{*}$, the desired foothold positions $\boldsymbol{r}_{F,k}^{*}$ and the measured feet positions $\boldsymbol{r}_{F,k}$ while in contact with the terrain for each foot and at every time-step. Secondly, we define the \textit{Foothold Tracking Score} (FTS) as the ratio of successfully tracked footholds over the total generated by the GP within an episode. At the end of each support phase, we check if feet which were previously in swing phase have contacted the ground within $\unit[5]{cm}$ of the target foothold in the $xy$ plane, and increment the FTS by one for each foot with a successful touchdown in the aforementioned region. These metrics are important with regard to the combined use of the GP and GC as they quantify how reliably a GC can execute the footholds generated by the GP. The planner has been trained to select footholds within a minimum distance of $\unit[5]{cm}$ from any changes in elevation exceeding $\unit[1]{cm}$. As long as the controller can maintain foothold tracking within this region, then the combined system is ensured to operate safely.

\paragraph*{Training Results} GC training endured for approximately 58 hours for \textit{Flat-World} and approximately 116 hours for \textit{Random-Stairs} and \textit{Temple-Ascent}. The discrepancy in durations is due to the increased computational cost incurred in the physics engine when evaluating contacts between the terrain mesh and the multi-body system. Training in \textit{Flat-World} results in a policy that succeeds in generalizing well to planar foothold tracking, while training in \textit{Random-Stairs} and \textit{Temple-Ascent} extends these capabilities to 3D. However, the stair-climbing agent trained in the latter case exhibits worse MER and FTS than those trained in the former. This difference is due to the difficulties in designing sampling schemes that always initialize the robot in valid initial states, but also to the similarity of the foothold targets generated by the GP as a result of the repetitive terrain features in the suite.

\begin{figure}[t]
    \centering
    \includegraphics[width=1.00\linewidth]{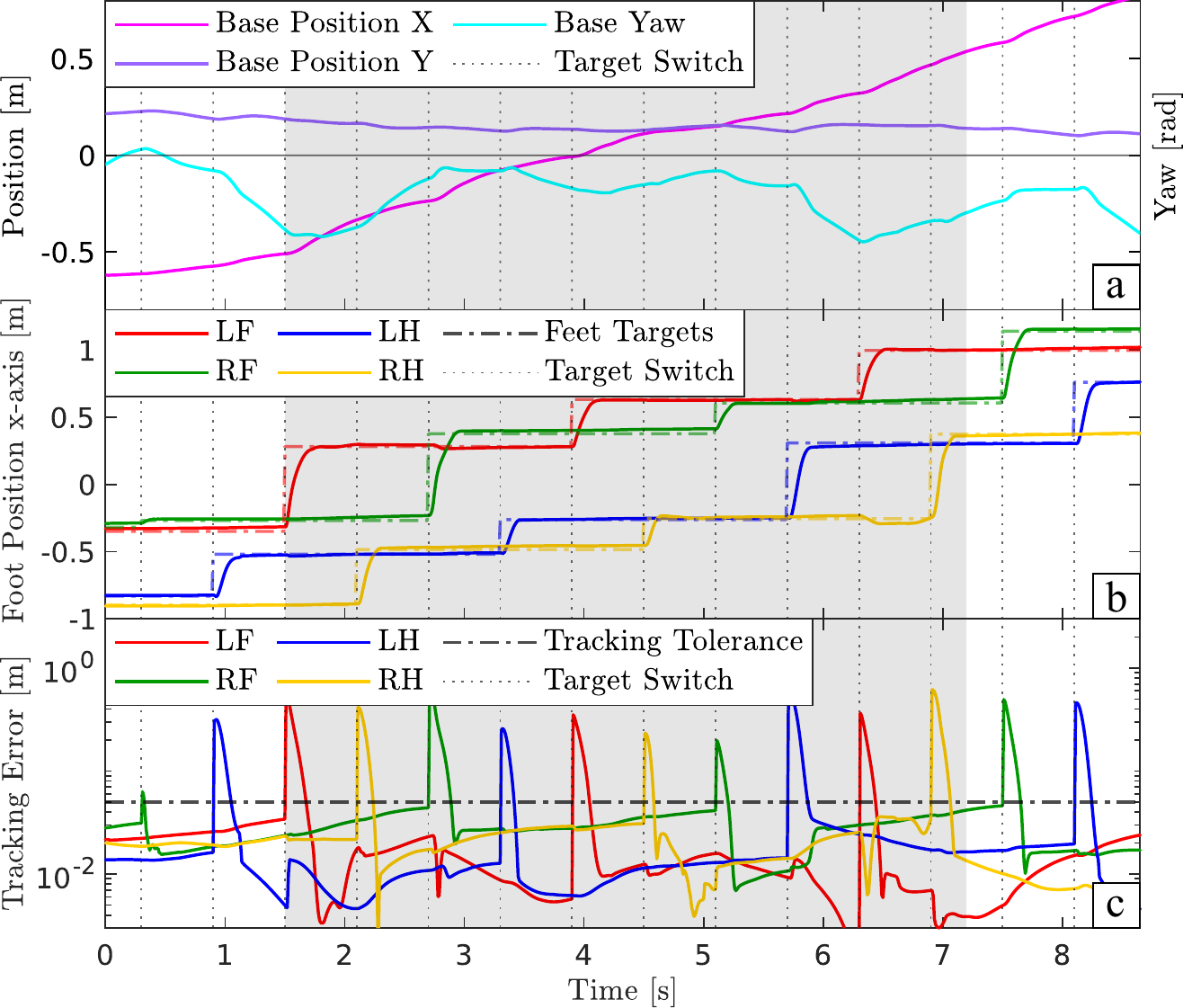}
    \caption{Plots of the GC policy overcoming the $\unit[40]{cm}$ gap: (b) planar pose of the base, (b) desired vs measured positions of the feet, and (c) log-scale norm of the foothold tracking errors in the X-Y plane. The vertical dotted lines denote the times at which the foothold targets change according to the phase plan, gray regions denote the time spent crossing the gap, and the horizontal line in (c) denotes the $\unit[5]{cm}$ error tolerance defined by the foothold tracking cylinder.}
    \label{fig:PerformanceOverGap}
    \vspace{-2mm}
\end{figure}

We evaluated the performance of the GC policies within \textit{Temple-Ascent} across five runs, each consisting of $100$ episodes with a maximum length of $\unit[90]{s}$. We also perturb the model of the robot (i.e., with which the GC was trained) to assess the robustness of the policies. Specifically, we increased the mass of the base by $\unit[25]{\%}$ and varied the lengths of the shank links by $\pm\unit[10]{\%}$. In each case, ESR, FTS, and FTER values were recorded in order to compute empirical means and standard deviations. The resulting performance measurements are presented in Tab.~\ref{table:Performance}\footnote{Although mean performance for stairs is $\geq\unit[90]{\%}$ in the nominal case, the variance is noticeably higher for the perturbed models, indicating that the policy is more sensitive w.r.t model variations than that for other terrains.} and Fig.~\ref{fig:PerformanceOverGap} shows time-series plots of the policy overcoming a large $\unit[40]{cm}$ gap. The final policies deployed are demonstrated in the supplementary video\footnote{\href{https://youtu.be/s1rrM1oczI4}{\url{https://youtu.be/s1rrM1oczI4}}}.

\subsection{Comparison to Existing Approaches}
\label{subsec:ComparisonToFreeGait}
As mentioned in Sec.~\ref{sec:Introduction}, most methods addressing planning and control of multi-contact motions employ optimization, sampling-based search, or a combination thereof. Although those relying solely on optimization~\cite{kuindersma2016optimization,winkler2018gait,fankhauser2018robust} can be kinodynamic, they are not feasible online, yet those that employ sampling-based search~\cite{tonneau2018efficient,tonneau2018reachability}, or both~\cite{Kalakrishnan:2011:LPC:1936801.1936805,doi:10.1002/rob.21610,griffin2019footstep} can be used online but remain kinostatic. The issue is that kinostatic methods are unable to fully exploit the dynamics of the system, and typically decouple foothold selection from the optimization of the base and feet motions, limiting their possible set of solutions. In addition, nearly all of the aforementioned approaches use some form of modelling~\cite{kuindersma2016optimization,Kalakrishnan:2011:LPC:1936801.1936805} or qualification~\cite{doi:10.1002/rob.21610,tonneau2018reachability,fankhauser2018robust} of the terrain. On the contrary, our approach relies on a minimal geometric representation of the terrain in the form of a height-map, performs gait-free kinodynamic planning by jointly optimizing base and swing feet motions together with the selection of footholds and is executable online.

\begin{figure}[t]
    \centering
    \includegraphics[width=1.0\linewidth,height=49mm]{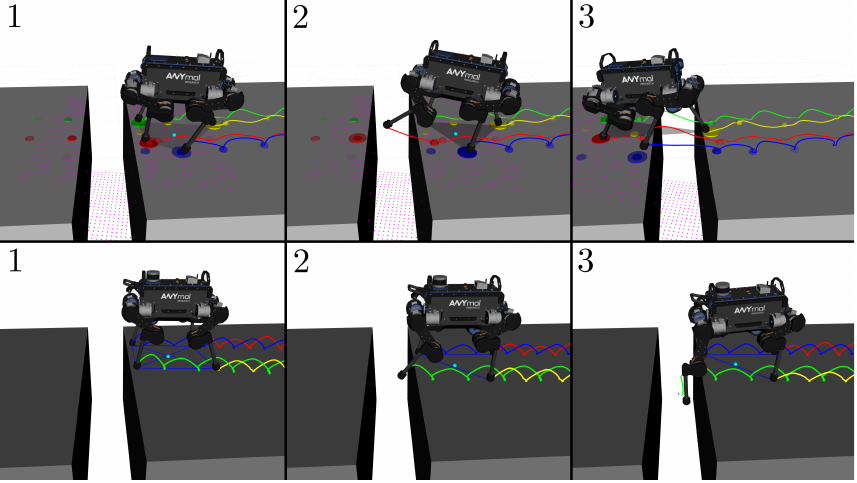}
    \caption{Snap-shots of the comparison between our policies (top row) and Free-Gait (bottom row) overcoming the $\unit[40]{cm}$ gap. The figures are numbered left-to-right to indicate the order of the frames.}
    \label{fig:ComparisonToFreeGait}
\end{figure}

Lastly, to illustrate how our method performs compared to others, we present an experiment in which the robot is to walk over a set of large gaps with different sizes. Specifically, we compare with \textit{Free-Gait}, a state-of-the-art model-based framework for perceptive locomotion~\cite{fankhauser2018robust} which jointly optimizes body poses and footholds using an Sequential Quadratic Programming solver. The experiment consists of two trials, corresponding to gaps of different sizes: $\unit[30, 40]{cm}$. Fig.~\ref{fig:ComparisonToFreeGait} provides snap-shots of the respective trials. Only our policies were able to traverse the $\unit[40]{cm}$ gap. Free-Gait relies on a heuristic scoring of the surrounding terrain and selects footholds using a deterministic greedy search within the vicinity of a nominal foothold. As it ensures that the CoM remains within some margin of the support polygon, it does not sacrifice static stability, even momentarily, to extend the reach of the front legs. Conversely, our planner can identify valid footholds across the gap in a single shot, and our controller then orientates the base to extend leg reach while retaining balance at all times.

\section{Conclusion \& Discussion}
\label{sec:Discussion}
This work proposes a new approach for training a two-layer hierarchy of NN policies which realizes terrain-aware locomotion for quadruped robots. We decompose locomotion into two parts trained independently using model-free DRL, and which interface via a carefully designed parameterization of quadrupedal gaits. As physical simulation incurs a high computational cost for use with DRL, our method reduces sample-complexity for training the gait planner by using a \textit{transition feasibility} criteria realized as convex LP to formulate the transition dynamics of an MDP. Within a certain context, our formulation can resemble bilevel optimization schemes used in nonlinear programming. The upper-level optimization performed by model-free DRL optimizes the selection of footholds and instantaneous CoM motion for the coincident terrain. The lower-level optimization performed by the model-based LP optimizes phase transitions, thus resulting in optimal CoM trajectories between the support phases proposed by the higher-level. Although in this work we only used a trivial cost for the LP, we could instead formulate a cost that penalizes contact forces and CoM accelerations, and include it as an additional reward term in the MDP. Such a coupling of the LP and MDP optimization objectives would allow us to reason about both feasibility and optimality of the resulting gait plan. Furthermore, as the planner learns to generate relatively conservative foothold sequences due to the restrained centroidal dynamics model, the burden of realizing fully dynamic and optimal motions for the base and swing-feet is placed on the gait controller. Lastly, we have demonstrated the efficacy of this approach by successfully training and testing in a suite of challenging terrain scenarios. The resulting policies not only prove to be effective in the suite of rigid non-flat terrains but also manage to generalize well to previously unseen cases.

\bibliographystyle{ieeetran}
\bibliography{main.bib}

\end{document}